# A Formal Framework for Speedup Learning
# from Problems and Solutions

**Prasad Tadepalli**                                                      TADEPALLI@CS.ORST.EDU
*Department of Computer Science*
*303 Dearborn Hall, Oregon State University*
*Corvallis, OR 97331*

**Balas K. Natarajan**                                                    NATARAJAN@HPL.HP.COM
*Hewlett Packard Research Labs*
*1501 Page Mill Road, Bldg 3U*
*Palo Alto, CA 94304*

## Abstract

Speedup learning seeks to improve the computational efficiency of problem solving with experience. In this paper, we develop a formal framework for learning efficient problem solving from random problems and their solutions. We apply this framework to two different representations of learned knowledge, namely control rules and macro-operators, and prove theorems that identify sufficient conditions for learning in each representation. Our proofs are constructive in that they are accompanied with learning algorithms. Our framework captures both empirical and explanation-based speedup learning in a unified fashion. We illustrate our framework with implementations in two domains: symbolic integration and Eight Puzzle. This work integrates many strands of experimental and theoretical work in machine learning, including empirical learning of control rules, macro-operator learning, Explanation-Based Learning (EBL), and Probably Approximately Correct (PAC) Learning.

## 1. Introduction

A lot of work in machine learning is in the context of concept learning. A prototypical example of this is learning to recognize hand-written characters from classified examples. Concept learning is the subject of an intense theoretical study under the name of "Probably Approximately Correct (PAC) Learning" – so called because the learner is required only to learn an approximation to the target concept with a high probability (Valiant, 1984). This rich and growing body of knowledge studies the possibility of learning approximations to concepts in different representations under various learning protocols. (See Natarajan, 1991, Anthony & Biggs, 1992, or Kearns & Vazirani, 1994 for a broad introduction.)

In this paper, we are concerned with a different kind of learning called "speedup learning," which deals with improving the computational efficiency of a problem solver with experience. One of the main differences between the concept learning and the speedup learning is that, in the latter, it is theoretically possible to solve the problems optimally using a brute-force problem solver. However, problem solving without learning is NP-hard in most of these domains, and hence is impractical in most cases. The role of learning can be seen as improving the efficiency of a brute-force problem solver by acquiring some "control knowledge" that is useful to guide problem solving in fruitful directions. In a concept learning task, before learning, there is not enough information to classify an example even by brute-force. Even though the speedup learning program has access to a brute-force problem solver, it is still a challenge to reformulate its knowledge in a way that makes problem solving efficient. There have been many successful speedup learning systems described in the experimental machine learning literature, PRODIGY (Minton, 1990) and SOAR (Laird, Rosenbloom, & Newell, 1986) being two of the most prominent ones.

Consider the domain of symbolic integration. Given the definition of the domain and a standard table of integrals, anyone has complete information on how to solve any solvable problem. Yet, while we humans are capable of solving problems of symbolic integration, we are by no means efficient in our methods. It appears that we need to examine sample instances, study solutions to these instances, and based on these solutions build up a set of heuristics that will enable us to quickly solve future problems. In this sense, the learning process has helped improve our computational efficiency.

We briefly describe the intuition behind our framework here, deferring the formal details to later sections. In essence, we would like our learning program to behave in the following manner: consider a class $\mathcal{M}$ of domains, such that each domain in the class is known to possess an efficient algorithm. We are interested in a learning algorithm for the class $\mathcal{M}$, an algorithm that takes as input the specification of a domain drawn from the class as well as sample instances of the problems in that domain and their solutions, and produces as output an efficient algorithm for the domain. As we will see, the sample instances and their solutions play a crucial role in the process, as in their absence, constructing an efficient problem solver for the input domain can be computationally intractable. In this paper, we are interested in examining the conditions under which such learning is made computationally efficient by using sample instances and solutions. We present a unified formal framework that captures both supervised and unsupervised forms of speedup learning, where examples of successful problem solving are provided by a teacher and by a search program respectively. Our framework is based on some of our previous work reported earlier (Natarajan & Tadepalli, 1988; Tadepalli, 1991a). Our methods of analysis are similar to that of PAC learning, in that we only require the learner to output an approximately correct problem solver with a high probability. Just as in PAC learning, we require the learner to be successful on any stationary problem distribution unknown to the learner.

There have been some other attempts to formalize speedup learning (e.g., Cohen, 1992, Greiner & Likuski, 1989, Subramanian & Hunter, 1992). However, most of these formalizations of speedup learning use a measure of problem-solving performance such as the number of nodes expanded in solving a problem (Cohen, 1992) or the number of unifications done in answering a query (Greiner & Likuski, 1989). We believe that these measures are too fine grained to be useful as a foundation for a robust theory of speedup learning comparable to the analysis of concept learning in the PAC-learning framework. Following the standard practice in complexity theory, we use the worst-case asymptotic complexity as our measure of performance. We require a successful speedup learning program to result in a problem solver whose worst-case asymptotic complexity is better than that of the nonlearning brute-force problem solver. Moreover, the learning itself must consume only a polynomial amount of time and a polynomial number of examples. Note that, according to our definition, the standard forms of compiler optimizations such as loop unrolling, and improvements in the hardware on which the program is run do not qualify as learning processes because they do not change the asymptotic complexity of problem solving. However, more sophisticated forms of program transformation such as partial evaluation are considered as learning provided they improve the asymptotic complexity of program execution. Thus, we believe that improving the worst-case asymptotic complexity of problem solving captures a cognitively interesting form of speedup learning. Although any decrease in asymptotic complexity is useful and interesting, in this paper we will be concerned with learning of polynomial-time problem solvers for domains which can only be solved in exponential time in the worst case without learning.

In Section 2, we introduce the preliminaries of problem solving and PAC learning. In Section 3, we introduce our formal framework for speedup learning. Drawing on prior results in PAC-learning, we prove a general theorem identifying conditions sufficient to allow such learning. In Section 4, we apply our framework to learning control rules and describe an implementation and experimental results in the symbolic integration domain. In Section 5, we apply our framework to learning macro-operators in the domain of Eight Puzzle. In Section 6, we discuss our work in relation to previous formalizations of speedup learning. In Section 7, we discuss some future extensions to our framework, including learning from unsolved problems and exercises. Section 8 concludes the paper.

## 2. Preliminaries

Without loss of generality, we assume $\Sigma = \{0, 1\}$ to be the alphabet of the language of state descriptions, and use $\Sigma^n$ for the set of binary strings of length $n$.

A *problem domain* $D$ is a tuple $\langle S, G, O \rangle$, where, $S = \Sigma^n$ is a set of *states*, $G$ is procedure to recognize a subset of states in $S$ as the *goal states*, and $O$ is a set of *operators* $\{o_1, \ldots, o_k\}$, where each $o_i$ is a procedure which takes a state in $S$ as input and outputs another state also in $S$. The combination of goals and operators is called the *specification* of $D$. A *meta-domain* $\mathcal{M}$ is a set of domains defined over the same set of states.

We denote the result of applying an operator $o$ to a state $s$ by $o(s)$. A *problem* is a state $s \in S$. A problem $s$ is *solvable* if there is a sequence of operators $\beta = (o_{x_1}, \ldots, o_{x_d})$, and a sequence of states $(s_0, \ldots, s_d)$, such that (a) $s = s_0$, (b) for all $i$ from 1 to $d$, $s_i = o_{x_i}(s_{i-1})$, and (c) $s_d$ satisfies the goal $G$. In this case, $\beta$ is a *solution sequence* of $s$, and $d$ is the *length* of the solution sequence $\beta$.

The *problem size* is a syntactic measure of the complexity of a problem such as its length when encoded in binary. If $s$ is an arbitrary problem in $S = \Sigma^n$, then its size $|s|$ is $n$.

Notice that our domain specification is not as explicit as the domain theory used in typical speedup learning programs like PRODIGY (Minton, 1990). The operators need not be described in the STRIPS formalism, and goals need not be logical formulas. In fact, they need not be declaratively represented at all, but may be described by procedures whose run time is reasonably bounded. Thus, our learning framework requires the learning techniques to be more independent of the operator representation than the traditional methods. This allows choosing the operator representation which is best suited to the domain rather than being constrained by the assumptions of the learning technique.

In the speedup learning systems studied by the experimental community, the goals and operators are usually parameterized. These systems also learn control rules and macro-operators with parameters. Learning parameterized rules and macro-operators makes it possible to apply them to problems of arbitrary size. Another advantage of parameterization is the ability to apply the same rule recursively many times, where each application binds the parameters to different arguments. Unfortunately, however, parameterization also increases the computational cost of instantiating the operators (or rules). When the number of parameters can be arbitrarily high, the instantiation problem is NP-complete in general. One way to theoretically limit this cost is to upper-bound the number of parameters of the operators, macro-operators, and control-rules to a constant. This ensures that both the time for instantiation and the number of different instantiations are polynomials in the length of the state description. It is possible to extend our results to such parameterized domains with some suitable restrictions on the number of parameters or their interdependencies (Tambe, Newell, & Rosenbloom, 1990). In fact, our application of the formal framework to the symbolic integration domain does involve an implicit parameter that denotes the subexpression of the current expression to which an operator is applied. However, for simplicity of exposition, we currently restrict our theoretical framework to nonparameterized operators.

A *problem solver* $f$ for a domain $D$ is a deterministic program that takes as input a problem, $s$, and outputs its solution sequence if such exists, or the special symbol "$\perp$" if it does not exist.

A *hypothesis space* $\mathcal{F}$ is a set of problem solvers. If $\mathcal{F}$ is a space of hypotheses, the restriction of $\mathcal{F}$ to problems of size $\leq n$ is called a *subspace* of hypotheses and is denoted by $\mathcal{F}_n$. Formally, for every $f \in \mathcal{F}$, there is a corresponding problem solver $f_n \in \mathcal{F}_n$ such that $f_n(x) = f(x)$ if $|x| \leq n$ and undefined otherwise. The *logarithmic dimension* or *l-dimension* of a hypothesis space $\mathcal{F}$ is defined to be $\log |\mathcal{F}_n|$ and is denoted by $dim(\mathcal{F}_n)$.

## 3. Learning from solved problems

In this section, we describe our learning framework. First, the domain specification is given to the learner. The teacher then selects an arbitrary problem distribution and a problem solver. We assume that there is at least one problem solver in the hypothesis space of the learner that is functionally equivalent to the teacher's problem solver, i.e., one which outputs the same solution as the teacher's problem solver on each problem. We call such a problem solver in the learner's hypothesis space, a *target problem solver*.

The learning algorithm has access to an oracle called SOLVED-PROBLEM. At each call, SOLVED-PROBLEM randomly chooses a problem in the current domain, solves it using the teacher's problem solver, and returns the ⟨*problem solution*⟩ pair, which is called an *example*. A *training sample* is a set of such training examples. We assume that if the problem is not solvable by the teacher's problem solver, it outputs the pair ⟨*problem*, $\perp$⟩.

Ideally, the goal of speedup learning is to find a target problem solver in the learner's hypothesis space. However, this is not always possible because our model of learning relies on randomly chosen training examples. Hence, we allow the learning algorithm to output an approximately correct problem solver with a high probability after seeing a reasonable number of examples. The problem solver needs only to be approximately correct in the sense that it may fail to produce a correct solution for a problem with a small probability even though the teacher succeeds in solving it. We are now ready to formally define our model of learning.

**Definition 1** *An algorithm $A$ is a speedup learning algorithm for a meta-domain $\mathcal{M}$ in a hypothesis space $\mathcal{F}$, if for any domain $D \in \mathcal{M}$, any choice of a problem distribution $P$, and any target problem solver $f \in \mathcal{F}$,*

1. *$A$ takes as input the specification of a domain $D \in \mathcal{M}$, maximum problem size $n$, an error parameter $\epsilon$, and a confidence parameter $\delta$;*

2. *$A$ may call SOLVED-PROBLEM, which returns examples $\langle x, f(x) \rangle$ for $D$, where $x$ is chosen with probability $P(x)$ from the problem set $\Sigma^n$; the number of oracle calls of $A$ must be polynomial in the maximum problem size $n$, $\frac{1}{\epsilon}$, $\frac{1}{\delta}$, and the length of its input; its running time must be polynomial in all the previous parameters and an upper bound $t$ on the running times of programs in $D$ on inputs of size $n$;*

3. *for all $D \in \mathcal{M}$ and all probability distributions $P$ over $\Sigma^n$, with probability at least $(1 - \delta)$, $A$ outputs a program $f'$ that approximates $\mathcal{F}$ in the sense that $\Sigma_{x \in \Delta} P(x) \leq \epsilon$, where $\Delta = \{x | f'(x) \neq f(x) \text{ and } f(x) \neq \perp\}$; and*

4. *there is a fixed polynomial $R$ such that, for a maximum problem size $n$, maximum solution length $L$, $\frac{1}{\epsilon}$, $\frac{1}{\delta}$, and the upper bound $t$ on the programs in $D$, if $A$ outputs $f'$, the run time of $f'$ is bounded by $R(n, L, t, \frac{1}{\epsilon}, \frac{1}{\delta})$.*

There are a few things that should be noted about this framework. Similar to the framework of Tadepalli (1991a), but unlike that of Natarajan and Tadepalli (1988), the learning algorithm is a function of the hypothesis space. Note that the teacher is free to generate solutions using any method. In particular, the teacher may be a human or a search program. The only requirement is that a target problem solver that is *functionally* equivalent to the teacher's problem solver exists in the learner's hypothesis space. This assumption is needed so that the learner can approximate the target problem solver arbitrarily closely by taking in more and more training examples. It would be impossible to do this if the target problem solver itself does not exist in the learner's hypothesis space.

Just as in the PAC-learning literature, learning must be successful independent of the choice of training distribution $P$. The problem solver $f'$ output by the learner is said to approximate the target problem solver, if they both produce the same solution with probability no less than $1 - \epsilon$, when tested on the problems sampled using $P$. Since the training problems are randomly chosen, they sometimes may not be representative, and the learner may fail to learn an approximately correct problem solver. Hence, we only require that such a problem solver is learned at least with a probability $1 - \delta$.

Finally, there is the requirement that the learned problem solver must be efficient. We capture this idea by insisting that it should run in time polynomial in various parameters, including the problem size, solution length, inverses of the error and reliability parameters $\epsilon$ and $\delta$, and the upper bound $t$ on the time needed for executing the domain operators. This last parameter $t$ factors out the time for executing individual operators from the problem-solving time, since this time is something that the learning algorithm cannot be expected to improve, because the operators are assumed to be opaque. In other words, we require only that the number of operator executions is polynomial in various parameters, even though the time for executing a single operator may be arbitrary but bounded.

The speedup achieved by the learner's problem solver is with respect to a default brute-force problem solver, which is the only one available to the learner before the learning begins, and *not* with respect to the problem solvers in the learner's hypothesis space. All the problem solvers in the hypothesis space of the learner are supposed to be efficient, since we are only measuring efficiency by coarse scales such as running in polynomial time. As we said earlier, we are not concerned with more refined notions of efficiency, such as improving the time complexity of problem solving from $O(n^3)$ to $O(n^2)$, in this paper.

Although we treated a problem solver as simply a deterministic program that maps problems to solutions, typically it consists of two components: a declarative representation of some kind of control knowledge (a function) that specifies which operator or operator sequence to apply in a given state, and an interpreter that uses the control knowledge to solve any problem in time polynomial in its size. Since the interpreter is usually fixed, the hypothesis space of problem solvers directly corresponds to a hypothesis space of possible control knowledge. Assuming that there is an efficient target problem solver in the hypothesis space of problem solvers implies that there is a target function in the corresponding hypothesis space of control knowledge. Speedup learning of a hypothesis space of target problem solvers can be achieved by PAC-learning of the corresponding hypothesis space of control knowledge. However, we do have an additional problem of converting problem-solution pairs of the target problem solver to examples of the target control knowledge. We take advantage of the domain specification (definition of goals and operators) in doing this conversion. Hence speedup learning in our framework consists of two steps: First, the problem-solution pairs of the target problem solver should be converted to examples of the target control knowledge using the domain specification. Second, the examples of target control knowledge must be generalized using some function learning scheme, and the result must be plugged into the interpreter to create an approximate problem solver.

For simplicity of exposition, this framework assumes that the maximum problem size $n$ is given. For a given problem distribution, this can also be easily estimated from examples by the standard procedure of starting with size 1 and iteratively doubling it and verifying it with a sufficiently large set of randomly generated problems (Natarajan, 1989).

**Definition 2** *A problem solver $f$ is consistent with a training sample if for every $\langle problem, solution \rangle$ pair in the training sample $f(problem) = solution$.*

Similar to many PAC-learning algorithms, the speedup learning algorithms we consider work by efficiently filtering the hypothesis space for a problem solver which is consistent with the training sample. Before we prove theorems about particular hypotheses spaces, we first state a general theorem which is a direct consequence of the results in PAC-learning of finite hypothesis spaces (Blumer, Ehrenfeucht, Haussler, & Warmuth, 1989).

**Theorem 1** *Let $\mathcal{M}$ be a meta-domain, and $\mathcal{F}$ be a hypothesis space of polynomial-time problem solvers for domains in $\mathcal{M}$. Let $dim(\mathcal{F}_n)$ be polynomially bounded in $n$. Then an algorithm is a speedup learning algorithm for $\mathcal{M}$ in $\mathcal{F}$, if it*

1. *takes the specification of $D \in \mathcal{M}$, $n$, $\epsilon$, and $\delta$ as inputs;*
2. *possibly calls the goals and operators in $D$;*
3. *collects $\frac{1}{\epsilon}(dim(\mathcal{F}_n) \ln 2 + \ln \frac{1}{\delta})$ training examples;*
4. *terminates in time polynomial in $n$, $\frac{1}{\epsilon}$, $\frac{1}{\delta}$, and in the sizes of the domain specification and the training examples; and*
5. *outputs a problem solver in $\mathcal{F}$ which is consistent with the training sample.*

In what follows, we refine this theorem to two particular hypothesis spaces: sets of control rules and macro-operators. We identify sufficient conditions to guarantee speedup learning in each of these two hypothesis spaces.

## 4. Learning control rules

One way to build efficient problem solvers is by learning control rules as in LEX (Mitchell, Utgoff, & Banerji, 1983) or in PRODIGY (Minton, 1990). Control rules reduce search by selecting, rejecting or ordering operators appropriately. In this section we consider learning of control rules that select appropriate operators to apply in a given state.

### 4.1 A theory of control-rule learning

A control rule is a pair $\langle U(o), o \rangle$, where $U(o)$ describes the set of problem states on which this rule selects the operator $o$. $U(o)$ is called the *select-set* of $o$.

We assume that the select-sets of operators of domains in $\mathcal{M}$ are described in some language $C$. We consider a hypothesis space $\mathcal{F}$ of problem solvers, where every problem solver consists of a set of select-sets in $C$, one for each operator in the domain. Let $C_n$ be the select-sets restricted to problems of size $\leq n$.

The hypothesis space $\mathcal{F}$ uses a fixed total ordering over the operators of the domain. This ordering is used to resolve conflicts between applicable operators when more than one select-set contains the given problem. In what follows, without loss of generality, we assume that the operators are numbered using this ordering. Given a problem and a set of control rules, a problem solver in $\mathcal{F}$ picks the least numbered operator whose select-set contains the problem, and applies it. This is repeated until the problem is solved or no select-set contains the current problem, in which case, the problem solver fails (see Figure 1). If the membership in the select-sets can be checked in polynomial time, then this problem solver runs in time polynomial in various parameters.

Now, we are ready to state and prove the main theorem of this section. The statement and proof of this theorem can be derived from previous results on learning sets with one-sided error (Natarajan, 1987). We prove it from the first principles for completeness.

Let $L$ denote a set of sentences, each of which represents a set of problems in the domain. There is a natural partial ordering over the elements of $L$ defined by the "more specific than" relation. A sentence is *more specific than* another if the set represented by the first sentence is a subset of that represented by the second sentence. We define a *most specific generalization* (MSG) of a set $S$ of problems in $L$ to be a sentence in $L$ which represents the most specific superset of $S$.

**Theorem 2** *A meta-domain $\mathcal{M}$ possesses a speedup learning algorithm in the hypothesis space $\mathcal{F}$ of problem solvers based on select-sets from $C$, if*

1. *every domain $D \in \mathcal{M}$ has a problem solver in $\mathcal{F}$ that solves any solvable problem in polynomial time;*

2. *for any set of problems in $S$, there is a unique most specific generalization in $C_n$, and it can be computed in polynomial time;*

3. *membership in the sets in $C_n$ can be checked in time polynomial in $n$; and*

4. *$\log |C_n|$ is a polynomial in $n$.*

```
procedure Control-rule-problem-solver
input x;
begin
    σ := "" ;
    while ¬Solved(x) do
      begin
              pick the least i s.t. x ∈ U(oᵢ);
              if no such i exists, halt with a failure;
              x := oᵢ(x);
              σ := Append(σ, oᵢ);
      end;
output σ;
end Control-rule-problem-solver
```

Figure 1: A problem solver that uses control rules

**Proof:** The key idea in the proof is as follows: Given a problem domain $D$, the learning algorithm will construct approximations to the select-sets of the operators of $D$ by finding the most specific generalizations of the example problems to which they are applied. If the conditions of the theorem are satisfied, this can be carried out in polynomial time, which is exponentially faster than the default brute-force search. With these select-sets in place, the algorithm **Control-rule-problem-solver** of Figure 1 behaves as an approximate problem solver for the domain $D$.

The rest of the proof deals with the details. Specifically, we will exhibit a speedup learning algorithm for $\mathcal{M}$. Let $D$ be a domain in $\mathcal{M}$.

Let $C$ be a language as in the statement of the theorem. By the conditions of the theorem, $C$ must possess an algorithm that finds the most specific generalization of a set of examples in polynomial time. The learning algorithm **Control-rule-learner** in Figure 2 uses this algorithm, called **Generalize**, to construct good approximations for the select-sets in $C$, and uses them to build a problem solver.

In particular, the **Control-rule-learner** works as follows. It first collects the required number of examples, and for each problem, obtains all its intermediate subproblems by applying its solution sequence to it. For each operator $o_i$ in the domain, it collects the set of subproblems for which it is the first operator applied in their solutions. It then calls **Generalize** on these sets $S(o_i)$, which outputs approximate select-sets $U(o_i)$.

We now show that the procedure **Control-rule-learner** of Figure 2 is indeed a learning algorithm for $\mathcal{M}$ in $\mathcal{F}$, if every domain in $\mathcal{M}$ has a problem solver in $\mathcal{F}$. First, we show that **Control-rule-learner** outputs a problem solver which is consistent with the training sample.

The proof is by induction on the length of the teacher's solutions of training problems. It is trivially true for any solutions of length 0. Assume that the above statement is true for any training problems and their intermediate subproblems which are solved in less than **procedure** Control-rule-learner
**input** $\epsilon, \delta, D = (G, O), n, L$
**begin**
    Let $O = \{o_i | i = 1, \ldots, k\}$;
    **initialize** $S(o_1), \ldots, S(o_k)$ to $\{\}$;

    /* **Section 1:** Generate examples for select-sets */

    **repeat** $\frac{1}{\epsilon}(k \log |C_n| \ln 2 + \ln \frac{1}{\delta})$ **times**
      **begin**
          **call** SOLVED-PROBLEM to obtain $(x, \sigma)$;
          If $\sigma \neq \perp$
          **then**
               Let $\sigma = $ "$o_{x_1}, o_{x_2}, \ldots, o_{x_r}$"
               $S(o_{x_1}) := S(o_{x_1}) \cup \{x\}$
               $S(o_{x_2}) := S(o_{x_2}) \cup \{o_{x_1}(x)\}$
               $\ldots$
               $S(o_{x_r}) := S(o_{x_r}) \cup \{o_{x_{r-1}}(\ldots(o_{x_1}(x))\ldots)\}$;
      **end**;

    /* **Section 2:** Construct approximations of select-sets */

    **for** $i := 1$ **through** $k$ **do**
          $U(o_i) := $ Generalize$(S(o_i))$;
    **output** the problem solver of Figure 1 with the learned $U(o)$'s
**end** Control-rule-learner

Figure 2: An algorithm for control rule learning

$r$ operator applications by the teacher. Consider a (sub)problem $x$ which is solved by the sequence $\sigma = $ "$o_{x_1}, o_{x_2}, ..., o_{x_r}$" by the teacher. The learning algorithm includes $x$ in the set $S(o_{x_1})$. When the learning algorithm generalizes this set, the learned select-set of $o_{x_1}$ includes $x$. Since all problem solvers in the hypothesis space including the target problem solver always use the least numbered operator whose select-set contains the problem, it follows that the target select-sets of operators $o_1, \ldots, o_{x_1-1}$ do not contain $x$. Moreover, since the learner finds the most specific generalization of the set of examples, the learned select-sets of operators $o_1, \ldots, o_{x_1-1}$ must be subsets of the corresponding target select-sets, and hence do not contain $x$. Hence $o_{x_1}$ is the least numbered operator whose select-set contains $x$ and will be selected by the learned problem solver to solve $x$. Since $o_{x_1}(x)$ is solved with a sequence of length less than $r$ by the teacher, by inductive hypothesis, it will be solved with the same sequence by the learned problem solver. Hence $x$ will be solved using $\sigma$ by the learned problem solver.

1. $\int k f(x) dx = k \int f(x) dx$
2. $\int f(x) - g(x) dx = \int f(x) dx - \int g(x) dx$
3. $\int f(x) + g(x) dx = \int f(x) dx + \int g(x) dx$
4. $\int f(x) * g(x) dx = g(x) \int f(x) dx - \int \{(Dg(x)x) \int f(x) dx\} dx$
5. $\int x^n dx = x^{(n+1)}/(n+1)$
6. $\int \sin x \, dx = (-\cos x)$
7. $\int \cos x \, dx = \sin x$

Figure 3: A table of integration operators

We now show that the sample size in our algorithm is sufficient for learning an approximate problem solver. For problems of size $n$ or less, each set $U(o)$ can be chosen in $|C_n|$ ways in Section 2 of the algorithm. Since there are $k$ operators, the number of distinct select-set tuples, and hence the number of distinct problem solvers that can be constructed in Section 2 is $|C_n|^k$. Hence $dim(\mathcal{F}_n) \leq k \log |C_n|$. Hence by Theorem 1 the sample size given in the algorithm is sufficient for learning.

Since $\log |C_n|$ is polynomial in $n$, if membership in the sets in $C_n$, and the most specific generalizations of the sets of examples can both be computed in polynomial time, then **Control-Rule-Learner** runs in polynomial time as well. Hence, by Theorem 1, it is a speedup learning algorithm for $\mathcal{M}$ in $\mathcal{F}$. $\square$

Note that the above theorem can also be stated using the on-line mistake-bound model, in which the learner incrementally updates a hypothesis whenever it cannot solve a new training problem in the same way as the teacher does, i.e., whenever the learner makes a "mistake" (Littlestone, 1988). This yields a slightly more general result than Theorem 2, because, under the same conditions of this theorem, the number of mistakes of the learner in the worst-case is polynomially bounded for any arbitrary choice of training examples, i.e., not necessarily generated using a fixed probability distribution. The mistake-bound algorithms can be converted to batch PAC-learning algorithms in a straightforward way (Littlestone, 1988).

## 4.2 Application to symbolic integration

We now consider an application of Theorem 2 to the domain of symbolic integration, as was done in the LEX program (Mitchell et al., 1983). We will show how this can be efficiently implemented using a straightforward application of Theorem 2 for a subset of LEX's domain.

Consider the class of symbolic integrals that can be solved by the standard integration operators. Let $\mathcal{M}$ be the set of domains whose operators are described by rules such as in Figure 3, and whose problems can be described by an unambiguous context free grammar $\Gamma$ such as shown in Figure 4.

Let $\alpha$ be any sentential form, i.e., a string of terminals and variables, of the grammar $\Gamma$ of Figure 4 derivable from the start symbol $Prob$. A sentential form $\alpha$ denotes the set of problems derivable from $\alpha$ using the productions of $\Gamma$. Consider a hypothesis space $\mathcal{F}$ of

$$Prob \rightarrow \int Exp \; d \; Var | DExpVar$$
$$Exp \rightarrow Term | Term + Exp | Term - Exp$$
$$Term \rightarrow P\text{-}term \; | \; P\text{-}term * Term | \; P\text{-}term \; / Term$$
$$P\text{-}term \rightarrow Const | Var | (-Term) | Trig | Power | Prob | (Exp)$$
$$Power \rightarrow (Var \uparrow Term)$$
$$Trig \rightarrow (\sin Var) | (\cos Var)$$
$$Const \rightarrow Int | a | k$$
$$Var \rightarrow x$$
$$Int \rightarrow 0 | 1 | 2 | 3 | 4 | 5 | 6 | 7 | 8 | 9$$

Figure 4: A grammar to generate the integration problems

problem solvers whose select-sets are represented by the sentential forms of the grammar $\Gamma$. We plan to show that the **Control-rule-learner**, with an appropriate **Generalize** routine that computes the MSG of a set of problems, is a learning algorithm for $\mathcal{M}$ in $\mathcal{F}$.

We first need a few definitions. A *parse tree* is an ordered tree where all nodes are labeled by the variables or terminals of the grammar, and the root is labeled by the start symbol. Moreover, if a node $V$ has children $V_1, \ldots, V_k$, then $V \rightarrow V_1, \ldots, V_k$ must be a production of the grammar. The string of symbols obtained by reading the leaves of the parse tree from left to right is called the *yield* of the tree (Hopcroft & Ullman, 1979). If the grammar is unambiguous, then for every sentence which can be generated by the grammar there is a unique parse tree which yields that sentence. This tree is called the parse of that sentence.

A *cap* of a tree $T$ is any ordered subtree $T'$ such that (a) all the nodes and edges of $T'$ are in $T$, (b) the root of $T$ is in $T'$, and (c) if a node is in $T'$, then its parent and its siblings in $T$ are also in $T'$.

Intuitively, a cap is obtained by pruning the subtrees rooted under some selected internal nodes in the parse tree and by making those nodes its leaves. Since the grammar is unambiguous, all the generalizations (in $C$) of an example correspond to the yields of various caps of the parse of that example. If there are two caps $c_1$ and $c_2$ for a parse tree such that $c_2$ is also a cap of $c_1$, then $c_1$'s yield is more specific than $c_2$'s, in that the set of sentences derivable from the yield of the former is a subset of the corresponding set derivable from the yield of the latter. We say that $c_1$ is more specific than $c_2$ in this case. $c_1$ is strictly more specific than $c_2$ if $c_1$ is more specific than $c_2$ and $c_1 \neq c_2$.

Given two or more parse trees for the same grammar, the *most specific cap* (MSC) is defined as a subtree which is a cap of all the parse trees such that no other common cap for these trees is strictly more specific. Since the caps of the parse tree of an example correspond to all possible generalizations of that example in our hypothesis space $C$, the yield of the MSC of the parse trees of a set of problems corresponds to the MSG of that set of examples.

We now describe the **Generalize** algorithm which computes the MSG of a set of examples by computing the MSC of their parse trees. The algorithm is to march down these parse trees simultaneously from the root, including a node and its siblings in the MSC if and only if they are all present in all the parse trees in exactly the same positions (and their parent is already included).

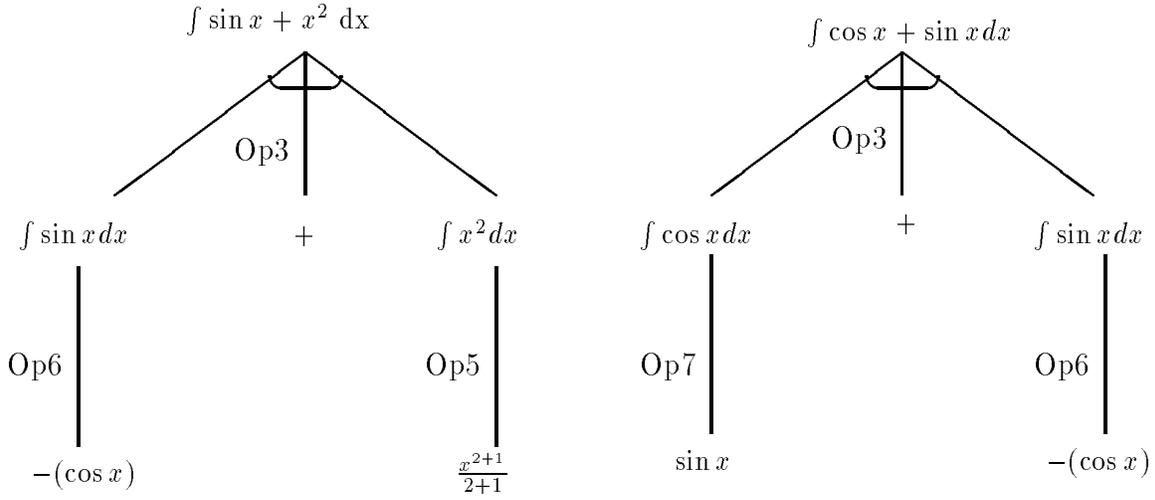

Figure 5: Tree representations of the solutions of the two examples

Consider, for example, that the program is given the following two examples. Figure 5 shows the solutions of the two problems in the form of trees.

1. $\int \sin x + x^2 dx \overset{op3}{\rightarrow} \int \sin x dx + \int x^2 dx \overset{op6}{\rightarrow} (-\cos x) + \int x^2 dx \overset{op5}{\rightarrow} (-\cos x) + \frac{x^{2+1}}{2+1}$

2. $\int \cos x + \sin x dx \overset{op3}{\rightarrow} \int \cos x dx + \int \sin x dx \overset{op7}{\rightarrow} \sin x + \int \sin x dx \overset{op6}{\rightarrow} \sin x + (-\cos x)$

From these two examples, the procedure **Control-rule-learner** generates the problem set $\{\int \sin x + x^2 dx, \int \cos x + \sin x dx\}$ for operator 3, and the singleton sets $\{\int x^2 dx\}$, $\{\int \sin x \, dx\}$, $\{\int \cos x dx\}$, for operators 5, 6, and 7 respectively. The MSGs of the singleton sets are the examples themselves. The parse trees of the two problems $\int \sin x + x^2 dx$ and $\int \cos x + \sin x dx$ for operator 3 are shown in Figure 6. The MSC of the two parse trees are marked with triangles. The yield of the MSC, $\int Trig + P\text{-}term \, dx$, corresponds to the unique MSG of the two examples.

**Generalize** computes the MSG of more than 2 examples incrementally by repeatedly finding the MSC of the parse trees of the current MSG (or the first problem) and the next problem. We are now ready to state and prove the following theorem.

**Theorem 3** *Let $C$ be the set of sentential forms derivable from the start symbol of an unambiguous context free grammar $\Gamma$. Let $\mathcal{F}$ be the hypothesis space of problem solvers defined using the select-sets from $C$. If each domain in the meta-domain $\mathcal{M}$ has a problem solver in $\mathcal{F}$ that correctly solves all solvable problems in that domain, then there is a speedup learning algorithm for $\mathcal{M}$ in $\mathcal{F}$.*

**Proof:** We show that **Control-rule-learner** is a learning algorithm for $\mathcal{M}$ in $\mathcal{F}$ by showing that the conditions of Theorem 2 hold. We already assumed the first condition of Theorem 2, namely the existence of complete problem solvers in $\mathcal{F}$.

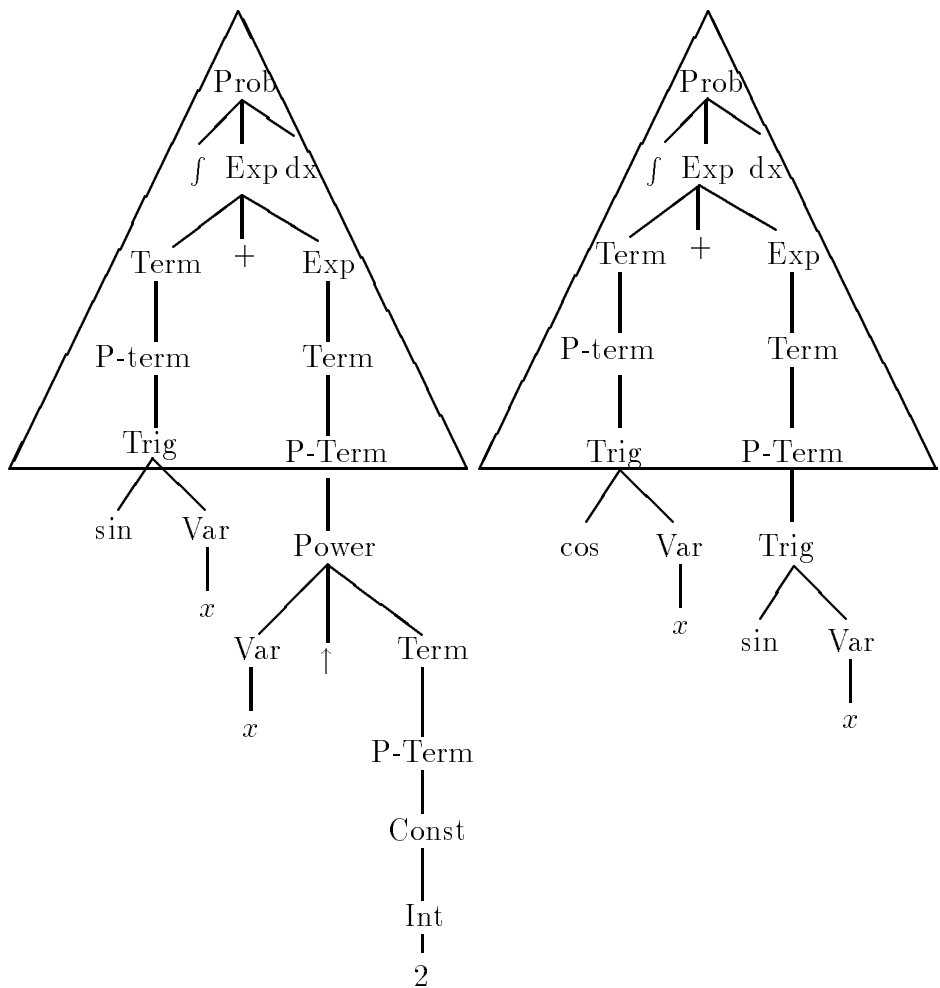

Figure 6: Finding the MSG of the two examples: $\int \sin x + x^2 dx$ and $\int \cos x + \sin x dx$

The MSG of a set of problems is unique because $\Gamma$ is unambiguous. As described earlier, the MSC can be computed in time linear in the number of examples and the sizes of the parse trees. Since parsing for unambiguous context free grammars can be done in time $O(n^2)$ (Earley, 1970), the MSG of a set of problems can be found in polynomial time. Thus the second condition of Theorem 2 is satisfied as well.

The third condition of Theorem 2 also holds since membership in select-sets corresponds to parsing which is an $O(n^2)$ problem. Finally, since the number of sentential forms of length $n$ is at most $c^n$ for some constant $c$, $\log |C_n|$ is of the order of $O(n)$, satisfying the last condition of Theorem 2. Hence it follows that $\mathcal{M}$ has a speedup learning algorithm in $\mathcal{F}$. $\square$

### 4.3 Experimental Results

The **Control-rule-learner** and the **Control-rule-problem-solver** are implemented in a program called SIMPLEX, and tested in the symbolic integration domain. The expressions to be solved are generated using the grammar in Figure 4. The domain has 39 operators including those in Figure 3, and some differentiation and simplification operators. Solving a problem consists of removing the integral sign and simplifying the result as much as possible. In this experiment, we assumed that the solutions to the problems are provided by an external teacher. To allow controlled experimentation with different training sets, we implemented the teacher using a set of select control rules. In fact, the teacher's control rules themselves were learned from 102 human-generated examples using the **Generalize** routine that computes the most specific generalization of a set of problems. Each such example consisted of a problem and the best first operator to apply on that problem. In this domain, each operator might be applicable to many parts of an expression. Hence the teacher-generated solution of a problem actually consists of a list of parameterized operators, where the parameter denotes the location (subexpression) at which it is applied. From these pairs, the program learns the select-sets as described in the previous section.

The problem solvers in the hypothesis space are assumed to employ post-order traversal of the expression tree to transform subexpressions by applying the operators. Since the learner's select-sets are obtained by most specific generalization of the training examples, and since both the learned problem solver and the target problem solver employ the same algorithm to traverse the expression tree, by the argument in the proof of Theorem 3, the learned problem solver is always consistent with the training examples. Hence the theorem is applicable to this parameterized domain as well.

We trained the system on integration problems that consisted of sums of products of powers of $x$ and some trigonometric functions of $x$. In particular, each problem was of the following form, where each coefficient was selected uniformly randomly from all its choices and independently from all other choices.[1]

$$\int \{0-9\} x^{\{3-9\}} + \{\sin x, \cos x, 0-9\} * x^2 + \{\sin x, \cos x, 0-9\} * x + \{\sin x, \cos x, 0-9\} dx$$

After each training example, the system was tested on a set of 100 test problems. The test problems were also selected using the same training distribution mentioned before. A test problem was counted as correctly solved by the learner if its solution exactly matches that of the teacher. This is a more conservative way of measuring accuracy than counting the problems which are reduced to equivalent expressions without the integral sign. It also forces the learner to simplify the results of integration in the same way that the teacher does.

Figure 7 shows the percentage of the test problems correctly solved from the test set averaged over 50 training trials plotted against the number of training examples. The error bars denote one standard deviation intervals on both sides of the mean. The learning converges quickly reaching 99% accuracy within 30 training examples. This is because each training example in fact gives raise to many small training experiences, each corresponding to one operator application.

---

1. We picked this narrow subset of the problems instead of the entire domain because (a) many problems in our domain do not have closed form solutions, and (b) since the teacher is to be first trained using hand-selected examples, it is tedious to do this on a large domain. The learning performance of the system is not sensitive to this choice of the problem distribution.

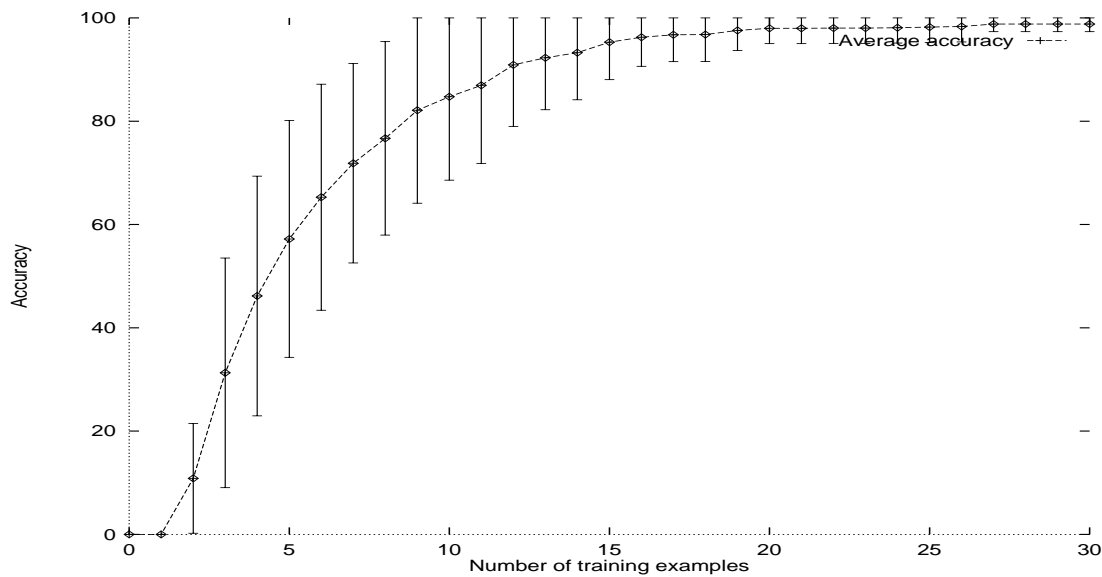

Figure 7: Learning curve for SIMPLEX; the error bars are one standard deviation away on either side of the mean.

## 5. Learning macro-operators

A macro-operator (or a macro) is any sequence of operators that achieves a subgoal. Macro-operators make the grain size of the search space coarser than the space of primitive operators, thereby increasing the efficiency of search. In this section, we consider the learning of macro-operators and formalize it using our speedup learning framework.

### 5.1 A theory of macro-operator learning

Here we make the assumption that states are representable as vectors of $n$ discrete valued features, where the maximum number of values a feature can take is bounded by a polynomial in $n$. In Rubik's Cube, the features are cubie (each of the 26 subcubes) names, and their values are cubie positions. In Eight Puzzle, the features are tiles, and their values are tile positions. We use the notation $\langle s^1, ..., s^n \rangle$ to represent a state $s$, where $s^i$ is the value of its $i^{th}$ feature.

A domain $D$ is *totally decomposable* if the effect of any operator in $D$ on a feature value is a function of the value of only that feature and is independent of all other feature values (Korf, 1985). Rubik's Cube is an example of a totally decomposable domain, because the effect of any turn on the position of a cubie is completely predictable from the original position of that cubie. Total decomposability is not obeyed by domains like Eight Puzzle. In Eight Puzzle, the effect of any operator like **up**, **down**, etc. on a tile depends not only on the position of that tile, but also on the position of the blank. Korf (1985) defined a more general notion of decomposability called "serial decomposability" which is applicable to such domains.

A domain is *serially decomposable* for a given total ordering on the set of features if the effect of any operator in the domain on a feature value is a function of the values of only that feature and all the features that precede it (Korf, 1985). If we treat the blank as a special feature in Eight Puzzle, then Eight Puzzle is serially decomposable for any feature ordering that orders the blank first. Note that serial decomposability is a property of the domain as well as its representation. If Eight Puzzle is represented with positions as features and tiles as their values, then it is not serially decomposable for any ordering of the features.

We assume that the goal is satisfied by a single goal state $g$ described by $\langle g^1, \ldots, g^n \rangle$. This assumption allows the learner to recognize when the subgoals are achieved. A domain satisfies *operator closure*, if the set of solvable states is closed under operators, i.e., every state reachable from a solvable state by an operator is solvable.

Consider a domain which is serially decomposable for some feature ordering $\Omega$. Without loss of generality, let $\Omega$ be the ordering $1, \ldots, n$. A macro table is a table of macros to achieve a single goal state, where the columns represent the features in the above ordering and the rows represent their possible values. A macro $M_{j,i}$ in the $i^{th}$ column of the $j^{th}$ row satisfies the *macro-table property* if whenever it is used in a solvable state $s$ where the features $1, \ldots, i-1$ have their goal values, $g^1, \ldots, g^{i-1}$, and the feature $i$ has the value $j$, the features 1 through $i$ in the resulting state are guaranteed to have the goal values $g^1$, ..., $g^i$. A macro $M_{j,i}$ is nonredundant if it satisfies the macro-table property and no strict prefix of $M_{j,i}$ satisfies it.

Korf showed that if a domain is serially decomposable and satisfies the operator closure, then it has a macro table (Korf, 1985). To see why, let $*$'s stand for some arbitrary (don't-care) feature values. For a domain which is serially decomposable with respect to $\Omega$, any operator sequence $\sigma$ that takes a state $\langle g^1, \ldots, g^{i-1}, j, *, \ldots, * \rangle$ to $\langle g^1, \ldots, g^i, *, \ldots, * \rangle$ satisfies the macro-table property, since the values of features 1 through $i$ in the latter state depend only on their values in the initial state, and not on the values of other features (represented with $*$'s). If the domain satisfies operator closure, there is bound to be some such operator sequence for any solvable state. Moreover, if $\sigma$ is redundant, $M_{j,i}$ can be replaced with its shortest prefix that satisfies the macro-table property. Hence, any serially decomposable domain that satisfies the operator closure has a macro table that contains only nonredundant macros. We call such a macro table nonredundant.

If a full macro table with appropriately ordered features is given, then it can be used to construct solutions from any initial state without any backtracking search as shown in Figure 8 (Korf, 1985). The features $i$ from 1 to $n$ are successively taken to their goal values, by applying macros $M_{j,i}$, where $j$ is the value of feature $i$ in the state before applying the macro. While the features 1 through $i-1$ may not have their goal values during the application of the macro, they all will regain their goal values along with the feature $i$ by the end of the application of the macro. Thus, any solvable problem is solved in $n$ macro applications by such problem solver.

**Definition 3** *A problem solver $f$ for a domain $D \in \mathcal{M}_\Omega$ is based on a macro-table $M$ if there is a total ordering $\Omega$ over the features such that,*

1. *$D$ is serially decomposable with respect to $\Omega$; and*

2. *$f$ constructs its solutions by running* Macro-problem-solver *on the macro table $M$ with the feature ordering $\Omega$.*

```
procedure Macro-problem-solver
input s          /* problem */
solution := "";
for i := 1 through n do
      begin
            j := sⁱ; /* = the value of the iᵗʰ feature of s */
            solution := Append(solution, M_{j,i})
            s := Apply(M_{j,i}, s);
      end;
output (solution);
end Macro-problem-solver
```

Figure 8: Korf's Macro Problem Solver

Korf's program fills the macro table by a single backward search from the goal state (Korf, 1985). In our implementation, macro-operators are learned incrementally by Iterative Deepening A* (IDA*) search. Given a random problem, the teacher constructs a solution as follows. It proceeds through the successive columns of the macro-table, starting with the first column. Before starting the search for a macro-operator in column $i$, the features 1 through $i-1$ are assumed to have the goal values already. If the value of the $i^{th}$ feature of the problem at hand is $j$, the teacher program seeks a macro-operator for the $j^{th}$ row and the $i^{th}$ column of the macro table. If there is already a previously learned macro in this location, the teacher simply applies it to the problem. Because the domain is serially decomposable with respect to the ordering $\Omega$, the features 1 through $i$ reach their goal values after this application. If there is no such previously learned macro-operator $M_{j,i}$, the teacher uses IDA* to search for an operator sequence that brings the features 1 through $i$ to their goal values. It applies this operator sequence to the current problem and proceeds to the next column. After going through all the columns of the macro-table in this manner, all the features would have reached their goal values. The entire operator sequence that transformed the initial state to the final state is returned as the solution.

It is important to note that the above implementation of the teacher is only one of many possibilities. The teacher oracle might use any other form of search or might be replaced by a human problem solver. The main requirement of our theory is only that there exists a target problem solver in the hypothesis space of the learner which is consistent with the teacher's problem solver. This requirement is fulfilled by the above implementation of the teacher because it reuses the macro-operators which are already learned by the learner whenever possible. This ensures that there is a single macro-table that can generate all solutions produced by the teacher. We describe a learning method called Serial Parsing, which works for any implementation of the teacher oracle as long as the above requirement is satisfied.

The Serial-parser (see Figure 9) uses teacher-given examples to incrementally build the macro table. To simplify the presentation, we assume that the program is given the number of features $n$, and the number of distinct feature values $v$, which together determine the problem size of our framework. Both of these can be estimated from examples at the cost of a little additional complexity of the learning algorithm.

**procedure** Serial-parser;
**input** $\epsilon, \delta, D = (G, O), n, v$;
**begin**
    Let $O = \{o_i | i = 1, \ldots, l\}$;
    **repeat** $m(\epsilon, \delta, n, v)$ **times**
        **begin**
            **call** SOLVED-PROBLEM to obtain $(x_0, \sigma)$;
            **if** $\sigma = \perp$ **then continue** with the next iteration ;
            Let $\sigma = $ "$o_{x_1}, o_{x_2}, \ldots, o_{x_r}$"
            /* Apply the operator sequence to the problem */
            **for** $k := 1$ **through** $r$ **do**
                $x_k := o_{x_k}(x_{k-1})$;
            /* Recognize the terminating points for macro-operators */
            Operator index $p := 0$;
            **for** $i := 1$ **through** $n$ **do**
                **begin**
                    $j := x_p^i$;     /* value of feature $i$ of $x_p$*/
                    Let $k \geq p$ be the smallest integer s.t. $\langle x_k^1, \ldots, x_k^i \rangle = \langle g^1, \ldots, g^i \rangle$
                    **if** $M_{j,i}$ is empty
                        **then** $M_{j,i} := $ "$o_{x_{p+1}}, o_{x_{p+2}}, \ldots, o_{x_k}$";
                    p := k;
                **end**
        **end**
    **output** macro table $M$
**end** Serial-parser

Figure 9: Serial Parsing Algorithm.

The idea behind the Serial-parser is simple. It collects a sufficient number of training problems and their solutions using SOLVED-PROBLEM. To each training problem $x_0$, it applies its solution sequence obtaining the sequence of intermediate states $x_1$ through $x_r$. Since it is known that the solutions to problems are generated using the macro problem solver with a known feature ordering, the solution sequence must be a composition of several macro-operators. It breaks this solution into its constituent macros $M_{j,i}$ for each feature $i$ by recognizing the earliest intermediate states in which the first $i$ features obtain their goal values. The macros are stored in the appropriate cells of the macro-table unless the cells have already been filled by previously learned macro-operators.

The result of this section can now be stated and proved.

**Theorem 4** Serial-parser *with* $m = \frac{1}{\epsilon}(nv \ln 2 + \ln \frac{1}{\delta})$ *training examples is a learning algorithm for* $\mathcal{M}_\Omega$ *in* $\mathcal{H}_\Omega$, *if*

1. *all domains in* $\mathcal{M}_\Omega$ *are serially decomposable with respect to* $\Omega$;

2. $\mathcal{H}_\Omega$ *is the set of all problem solvers based on nonredundant macro-tables with the feature ordering* $\Omega$ *for domains in* $\mathcal{M}_\Omega$; *and*

3. *the number of distinct feature values* $v$ *is bounded by a polynomial function of maximum problem size* $n$.

**Proof:** Without loss of generality, assume that $\Omega$ orders the features in the increasing order.

First, from Figure 8, we observe that the **Macro-problem-solver** runs in time $O(ntl)$ assuming that the time to apply a single operator is bounded by $t$ and the maximum length of a macro in the macro table is bounded by $l$. Hence $\mathcal{H}_\Omega$ is a set of polynomial-time problem solvers.

The **Serial-parser** stores the shortest operator subsequence that occurs between any state $\langle g^1, \ldots, g^{i-1}, j, *, \ldots, * \rangle$, and $\langle g^1, \ldots, g^{i-1}, g^i, *, \ldots, * \rangle$ as the macro $M_{j,i}$. Since the domain is serially decomposable for the feature ordering $1, \ldots, n$, the effect of any operator or macro-operator on features 1 through $i$ is not dependent on the values of features greater than $i$. Hence, it satisfies the macro-table property. In fact, $M_{j,i}$ must be identical to the corresponding macro in the target problem solver's macro table. This is so because the target problem solver is based on a nonredundant macro table, and any subsequence of the solution which has $M_{j,i}$ as a strict prefix would be redundant. Since all the macros present in the solutions of the training problems are thus correctly extracted by the **Serial-parser**, the **Macro-problem-solver** will be able to reproduce the solutions of all these problems using the learned macro table. Hence the problem solver output by the learning algorithm is consistent with the training sample.

Since the macros extracted by the **Serial-parser** always match the corresponding macros in the target macro table, the only way in which the learned macro table might fail to produce a solution given by the teacher is when some necessary macro-operator in the target macro table has never been learned. Since there are $n$ rows and $v$ columns, there are at most $nv$ macros, and any subset of these macros could be missing in a learned macro table. Hence the number of different macro tables or problem solvers that can be learned from a given target macro table is upper-bounded by $2^{nv}$. This is the effective hypothesis space of the learner. Hence, by Theorem 1, $m = \frac{1}{\epsilon}(nv \ln 2 + \ln \frac{1}{\delta})$ examples is sufficient to ensure learnability.

It is easy to see from Figure 9 that the running time of **Serial-parser** is bounded by $O(mrnt)$, where $r$ is the length of longest solution in the training sample, and the other parameters are as defined above. Since $m = O(\frac{nv}{\epsilon} \ln \frac{1}{\delta})$, the run time of **Serial-parser** is polynomial in all the required parameters. Hence it is a speedup learning algorithm for $\mathcal{M}_\Omega$ in $\mathcal{H}_\Omega$. $\square$

The above theorem shows that the **Serial-parser** exploits serial decomposability, a problem-space structure which allows it to compress the potentially exponential number of solutions into a polynomial size macro table. Serial Parsing requires that the teacher's solutions (provided by the SOLVED-PROBLEM) can be derived using a single problem solver, i.e., a single macro table. We satisfied this requirement by letting the teacher search for a macro-operator to solve a subgoal only when such macro-operator has not been previously learned. This ensures that there is always a macro table in the learner's search space which is consistent with all the solutions generated thus far. This approach closely integrates the "learner" and the "teacher" and brings our system closer to the previous implementations of unsupervised speedup learning such as SOAR (Laird et al., 1986).

To see the importance of the above requirement, consider what happens if the teacher uses some form of admissible search algorithm to give an optimal solution to every Eight Puzzle problem it is asked to solve. If the problems are chosen uniformly randomly, it is highly unlikely that all these optimal solutions can be derived from any single macro-table. The difficulty of finding optimal solutions for the $N \times N$ generalization of Eight Puzzle argues even more strongly against that possibility for bigger puzzles (Ratner & Warmuth, 1986). This suggests that the teacher is not free to use any problem solving or search method to solve problems, if the learning has to be successful. However, if the learner is allowed to ask queries, i.e., ask the teacher to solve carefully designed problems, the situation is different. Then the learner can ask the teacher to solve a problem designed specifically so that its solution would fit a particular cell in the macro-table. In fact, in our experiment described in the next section, the teacher uses search only to solve the subproblems that correspond to individual cells in the macro-table. Instead of interpreting this as the teacher ensuring that there is a single macro-table which is consistent with all its solutions, we can think of the teacher to be just the search program which solves the subproblems. Given a problem, the learner decomposes it into subproblems and tries to use the already learned macro-operators in its macro-table to solve them. Whenever a particular subproblem does not have a corresponding macro-operator in its table, it simply calls the "teacher" to solve it by search and stores the solution in its table. This is analogous to asking membership queries in one of Angluin's models of PAC learning (Angluin, 1988). With this membership query model, it is no longer required that there is a problem solver in the learner's hypothesis space which is consistent with the teacher's. There is also no guarantee that the learner and the teacher produce the same solutions on random problems. In fact, this is most likely not the case, because the learner uses the macro-table to produce its solutions and the teacher may not. For example, if the teacher always finds the shortest solution to a problem by search, then its solutions to randomly chosen problems are likely to be shorter than those produced by the learner.

## 5.2 Experimental Results

In this section, we illustrate an application of the theory to the Eight Puzzle domain.

In Eight Puzzle, let r, l, u, and d represent the primitive operators of moving a tile right, left, up, and down respectively. Macros are represented as strings made up of these letters. For example, the string "dr" represents down followed by right. For notational ease, features (tiles) are labeled from 0 to 8, 0 standing for the blank and $i$ for tile $i$. From the argument of the previous section, it is serially decomposable for the feature ordering 0 through 8. A macro $M_{j,i}$ represents the sequence of moves needed to get the $i^{th}$ tile to the goal position from its current position $j$, while preserving the positions of all previous tiles including the blank.

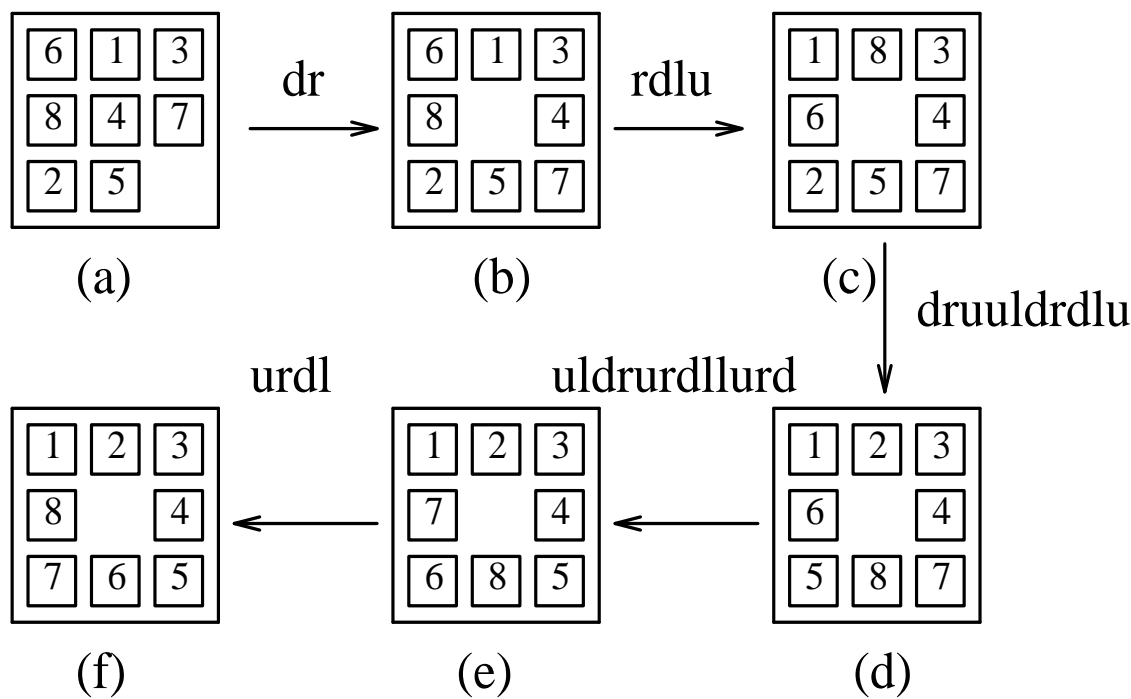

Figure 10: An Eight Puzzle problem and the intermediate subgoals. The macro-operators solve the successive subgoals.

The teacher is implemented as described in the previous section. If there is an applicable macro-operator which is previously learned to solve the next subgoal, it uses it. Otherwise, it uses IDA* with the Manhattan-distance heuristic. Since the teacher program is search-based and uses the current macro-table of the learner, our implementation actually blurs the distinction between the teacher and the learner, and more closely approximates the speedup learning systems such as SOAR and PRODIGY which learn in unsupervised mode, i.e., without being given the solutions.[2]

The board position (a) in Figure 10 represents the initial state and the board position (f) represents the goal state. The tile positions are numbered by the tile numbers in the goal state, which are assumed to be fixed. Hence the top left tile position is numbered 1, top middle tile position is numbered 2, and so on. The center tile position is numbered 0. Given the problem in the figure, our program looks for a macro-operator that solves the first subgoal. If the given feature ordering is 0 through 8, it first tries to take the blank from position 5 to its goal position, 0. Assuming that there is already a macro-operator $M_{5,0} = $ "dr" learned, it uses that macro, reaching the position (c) in the figure. It then looks for a macro-operator that takes tile 1 and the blank to their goal positions. Assuming that this macro-operator has not already been learned, it uses IDA* and finds that the macro-operator "rdlu" is appropriate in that it brings both the tile 1 and the blank to their goal positions. It applies this macro-operator, reaching the board position (b). It stores this macro in its table as $M_{2,1}$, since the position of tile 1 in board (b) is 2. It proceeds similarly through tiles 2 to 8, bringing them into their goal positions and learning new macros when needed. Note that since tiles 3 and 4 both have reached their goal positions along with tile 2 in board (d), a null macro will be stored in the corresponding cells of the macro-table, $M_{3,3}$ and $M_{4,4}$ respectively. Thus, from this example alone, the program can potentially learn 7 macros including the null macros.

---

2. In fact, to make our implementation cleaner and more efficient, the teacher, rather than giving the entire solution to a problem to the learner which it should then parse into macro-operators, directly gives it new macro-operators found by search. They are simply stored by the learner in its macro-table. This avoids the serial parsing step which would be needed if the teacher program is opaque.

The program was trained with 40 training examples using a fixed macro-table and tested after each 2 training examples on a sample of 100 random test examples. The training and test examples were selected using the uniform distribution over all solvable problems. The results in Figure 11 are averages over 50 different training sets. The error bars denote one standard deviation intervals on both sides of the mean. The learning converges quickly reaching a 98.7% average accuracy within 40 training examples. As can be expected, this is much smaller than the worst-case theoretical bound of 585 examples with $\delta = 0.1$ and 90% accuracy ($\epsilon = 0.1$). Knowing that there are only 35 nontrivial macros in the Eight-Puzzle domain reduces the theoretical bound to 266, which is still much higher than the examples needed in practice.

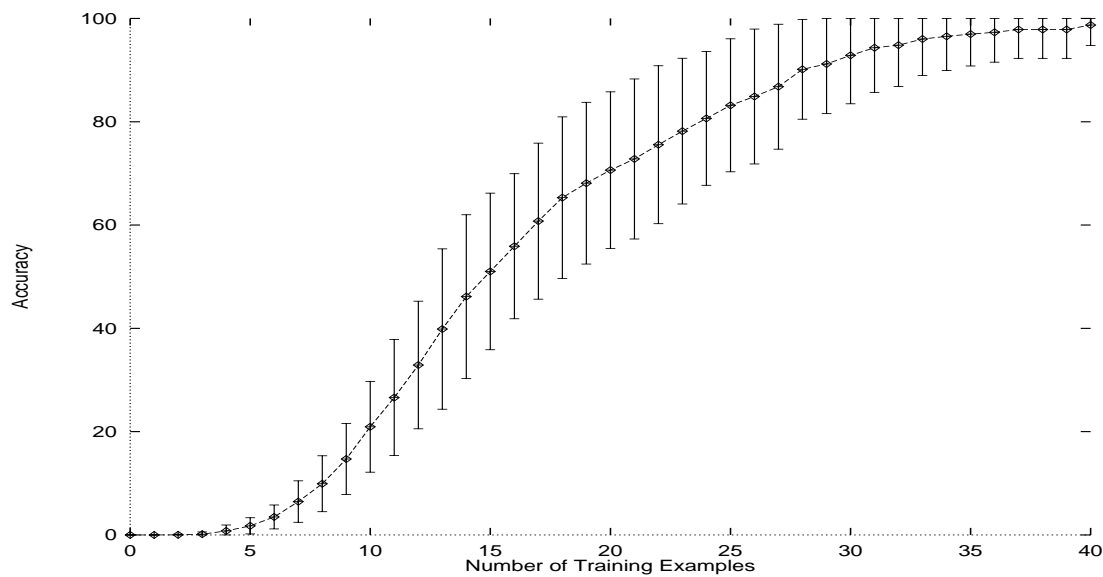

Figure 11: Learning curve for the macro-operator learning program; the error bars are one standard deviation away on both sides of the mean.

The sigmoid shape of the learning curve is worth noting. In the beginning, the learning is slow because solving a new test problem requires correct macro-operators for *all* the subgoals in its solution. With only a limited experience, it is likely that one or more of these are missing, which means that the problem cannot be solved. But with training, the effect of multiple learning experiences that correspond to the different subgoals in each training example enhances the learning speed, leading to a steep increase in the performance.

## 6. Discussion and Related Work

Recently, there have been a few formal frameworks proposed to capture speedup learning. For example, Cohen (1992) analyzes a "Solution Path Caching" mechanism and shows that organizing the solutions of the training problems in a tree and restricting the search of the problem solver to this tree improves the performance of the problem solver in the sense of reducing the number of nodes searched with a high probability. However, Cohen's results do not guarantee an eventual convergence to an *efficient* problem solver, but only to an optimal problem solver achievable by restricting the search to the tree of solutions of the training problems. By defining learning as producing a polynomial-time problem solver as opposed to simply running faster than the original problem solver, we have more stringent conditions on successful learning in our framework. For example, in domains like the Eight Puzzle, Solution Path Caching will produce an exponentially large tree of solutions, since each solution generated by the macro-table is stored as a path in the tree. Learning such large trees will need exponentially large number of examples and exponentially long running time. In retrospect, this is not surprising because solution path caching is a weak learning method that does not assume or exploit any structure in the problem space. Either a domain has some structure and hence significant speedup is achievable by exploiting it in some learning algorithm, or it does not have any structure, in which case learning can only have limited benefit. We believe that the role of a theory of speedup learning is to distinguish between these two cases, and provide learning algorithms for cases in which significant speedups are achievable. The validity of this general methodology is already borne out by the rich body of results in computational learning theory literature in the context of concept learning. Our aim is to transfer this methodology to speedup learning, identify problem domains for which effective speedup is possible, and build speedup learning algorithms for them.

Our work was originally aimed at formalizing a form of Explanation-Based Learning (EBL) (Tadepalli, 1991a). EBL constructs a proof of how a problem is solved in the training example using an explicit form of domain theory, and then generalizes and transforms that proof to a control rule or a macro-operator, which is justified by the original domain theory (Mitchell, Keller, & Kedar-Cabelli, 1986; DeJong & Mooney, 1986; Minton, 1990; Shavlik, 1990). Like Solution Path Caching, EBL is a weak learning method, and in general, cannot be expected to improve the performance. Indeed, the results in the speedup learning literature suggest that EBL could lead to problem solvers which are much more *inefficient* than the original problem solvers (Minton, 1990; Etzioni, 1993). However, depending on the structure of the problem space used, and the way in which EBL's domain theory is coded and used, it is possible for EBL to learn successfully in some situations. For example, Etzioni showed that in the PRODIGY system, EBL's success hinges on its ability to find constant-size nonrecursive proofs that show that choosing some operators in certain states is always bad (or always good) (Etzioni, 1993). Such constant-size proofs result in constant-size control rules, which are inexpensive to match. If there is a finite set of such control rules that can reduce the number of states expanded in problem solving from an exponential function of the state size to a polynomial function, the problem solving can be guaranteed to take only polynomial time (Etzioni, 1993, Proposition 2.1., pg. 102). Etzioni's original system STATIC exploited this structural feature of the problem space to learn efficient problem solvers without using any examples (Etzioni, 1993). A subsequent system called

DYNAMIC used examples to identify the problems to be explained (Perez & Etzioni, 1992). The examples play two roles in our theory: first, they provide distribution information that determines which macro-operators or control rules are worth learning, and second, they help the learner avoid expensive search for solutions. Perez and Etzioni (1992) separate these two roles, and use examples only to learn the distribution information.

While the conclusions of Etzioni (1993) may be read as too pessimistic for rules derived from recursive explanations, our results with macro-tables show that Explanation-Based Learning can be used to learn efficient polynomial-time problem solvers for arbitrary problem distributions, if the domain exhibits some structure such as serial decomposability. The application of our theory to learning macro-operators can be used to explain the success of SOAR in domains like Eight Puzzle (Laird et al., 1986). (The version of EBL used in SOAR is called Chunking.) Recall that Serial Parsing is given the order in which the subgoals are achieved. In systems like SOAR that successfully learn macros using EBL, the goal ordering is implicitly given by defining the subgoals such that they are successively inclusive (Laird et al., 1986). For example, in Eight Puzzle, the goals are "getting the blank in correct position," "getting the blank and tile 1 in correct positions," "getting the blank and tiles 1 and 2 in correct positions," and so on. This representational trick combined with the serial decomposability of the domain is mainly responsible for SOAR's success in learning macro-operators in Eight Puzzle.

The differences between Serial Parsing and EBL/chunking algorithms are worth considering. Unlike the operators used in EBL, the operators of Serial Parsing are opaque, and are not inspectable by the learning mechanism. To some extent, SOAR's operators are also opaque to its learning method in that the learning mechanism has only knowledge of which objects are "touched" by the operators, but does not have access to the operators themselves (Laird et al., 1986).[3] This suggests that, unlike in EBL (Mitchell et al., 1986), it is not necessary to have access to declaratively represented operators to achieve speedup using macro-operators. Knowing the feature ordering which makes the domain serially decomposable is sufficient to infer the appropriate conditions to apply a macro-operator, which is the main goal of chunking or the EBL process. If such a feature ordering is not known, neither EBL nor chunking might converge with a small number of macro-operators without some kind of utility analysis (Minton, 1990).

Tadepalli (1991b) describes a method called Batch Parsing, which learns the correct feature ordering along with the macro table. The basic idea here is to learn the macro table column by column, using multiple examples to disambiguate the feature that corresponds to a given column. While this method works without backtracking for Eight Puzzle, it is possible to construct domains for which it gets misled into wrong feature choice, and needs to backtrack.[4] It is not known whether there is a provably correct speedup learning algorithm that learns a correct feature ordering and the macro-table, from examples of solutions constructed from that macro-table. Bylander (1992) shows that detecting serial decomposability without examples is NP-hard in general. The existence of macro-tables is only guaranteed if there is a unique goal state and the operator closure is satisfied. Checking these properties is, in general, PSPACE-hard (Bylander, 1992). However, it may be easier to check these properties under some conditions. For example, Chalasani et al. (1991) describe an algorithm that detects serial and total decomposability for permutation groups (Chalasani, Etzioni, & Mount, 1991). If the operators are defined in STRIPS notation, it may sometimes be possible to check sufficient conditions for serial decomposability by constructing a graph of dependencies among the domain features and checking that it has no cycles. Similarly a sufficient condition for operator closure is that every operator has an inverse, which may be possible to check if we have access to explicit definitions of operators.

---

3. SOAR also makes the macro-table method applicable to any goal using another representational trick, i.e., by parameterizing the tiles rather than labeling them with fixed numbers.

4. We thank Prasad Chalasani for illustrating this.

There is a lot of interesting theoretical work in the area of speeding up logic programs. Greiner and Likuski (1989) introduced a model of speedup learning where redundant macro-rules are added to a base-level domain theory of Horn-rules (Greiner & Likuski, 1989; Greiner, 1991). Subramanian and Hunter (1992) extended this work by developing fine-grained cost models for theorem proving in recursive Horn-theories and using them to derive "utility theorems" that describe the conditions under which such redundant macro-rule learning is beneficial. Greiner and Jurisica (1992) describe a method called PALO that is based on hill climbing over a set of optimization transformations on the problem solver. Each transformation is only made if it significantly improves the problem solver's performance on a randomly chosen set of training problems. The program is guaranteed to converge to an approximate locally optimal problem solver with a high probability. The work of Gratch and DeJong (1992) in the COMPOSER system follows a similar strategy of applying a series of transformations which are proved useful on a training sample until the performance no longer improves.

One difference between our approach and all these methods is that our cost model is much more coarse than the others. In particular, we only require that the output problem solver must run in polynomial time, while the previous works have more fine-grained cost models. An advantage of the fine-grained models is that they could make more precise predictions. However, one also needs to know a lot more information such as problem distributions to make these predictions. In contrast, our goal is to identify structure in the problem space that guarantees qualitatively significant speedup with a reasonably small amount of training. In other words, we are seeking robust results which may not be as fine grained, but exploit interesting problem-space structure, and are amenable to coarse theoretical analysis.

There is a lot of scope for combining these two kinds of models, however. A coarse model may be used to make a quick and dirty analysis of the domain and identify possible optimizations and resulting speedups, and a detailed model may then be used to fine-tune the optimizations. For example, one of the interesting theorems proved by Subramanian and Hunter (1992) is that even adding a single redundant macro-rule which can be proved or disproved in a constant time can increase the overall cost of theorem proving exponentially! The reason for this is that macro-rules increase the number of different ways a goal may be proved, and increase the branching factor of search. Even a small increase in branching factor from 1 to 2 could exponentially increase the theorem proving cost for some problem distributions. The authors identify a condition called "separability," which, if preserved by the macro-rule learner, will have no negative impact and might have exponential benefit. Simply stated, separability exists when exactly one choice of rule is explored at every node in every computation, and hence corresponds to backtrack-free search. Not surprisingly, our problem solvers which are based on control rules and macro-operators also rely on backtrack-free search for efficiency. In fact, Subramanian and Hunter (1992) present an example where adding a single redundant macro-rule creates separability of the sort present in our macro-tables, thereby exponentially speeding up problem solving. It would be interesting to explore ways of transforming domain theories in a way that separability is preserved or created.

Our framework captures both empirical and explanation-based speedup learning methods in a uniform manner. Our SIMPLEX system is designed after LEX(1), which is described as an "empirical learning system" (Mitchell et al., 1983), and our macro-table learner is similar to SOAR, which is described as an "explanation-based learner" (Laird et al., 1986). We view the speedup learning problem as one of finding a close approximation of the target problem solver from examples of that problem solver and the domain specification by efficiently searching the hypothesis space of problem solvers. Generally there are two kinds of constraints obeyed by the problem solvers in the hypothesis space. One kind are the semantic constraints which are obeyed by all domains in the meta-domain. For example, serial decomposability is such a constraint. The other kind are syntactic constraints on the structure of the target problem solver. For example, the constraints that the target problem solver is organized as a macro-table or as a set of control-rules with left hand sides which are sentential forms of a grammar are examples of syntactic constraints. The syntactic and semantic constraints on the target problem solver help bias the learner, in that they improve its ability to generalize from a small number of training examples. Generally speaking, the semantic constraint is stronger in EBL systems and the syntactic constraint is stronger in empirical learning systems. Depending on the structure of these two kinds of constraints, the learner may adopt a variety of search strategies to find an approximation to the target problem solver in the hypothesis space. In general, all speedup learning systems assume that their representational structures — macros, control rules, or whatever else — are adequate to succinctly represent the control knowledge needed to efficiently solve the problems in their domain. In addition to syntactic and semantic biases which restrict the hypothesis space of problem solvers, a learning system might also incorporate preference biases, for example, prefer shorter rules, or rules derived from shorter explanations. Bias specifies the conditions under which learning succeeds and also provides the justification for the learning algorithm.

Speedup learning systems sometimes suffer from what has been called the "utility problem," which is the inefficiency of the learned problem solver caused by the proliferation of learned control knowledge which is too expensive to use (Minton, 1990). Our approach suggests that the utility problem can be solved in some cases by constraining the target problem solver so that it only learns efficient forms of control knowledge (properly indexed macro-operators or control rules) and uses them in a controlled fashion. Since the utility problem is unsolvable in general (Minton, 1990), our approach suggests a way to identify the cases in which it *can* be solved and precisely characterize them.

Khardon (1996) extends our work to the reinforcement learning problem where the goal is to learn an efficient action strategy, i.e., a mapping from sensory inputs to actions, in a stochastic domain. Unlike the typical reinforcement learning algorithms where action strategies are learned indirectly by learning value functions over states or state-action pairs (Kaelbling, Littman, & Moore, 1996), here the approach is to learn them directly by empirical generalization of action sequences observed from a knowledgeable teacher. Khardon shows that action strategies represented as systems of parameterized production rules with small preconditions are efficiently learnable in this framework using a greedy algorithm similar to that of Rivest (1987). Also, unlike in reinforcement learning, the goal is to closely approximate the teacher's action strategy rather than to learn the optimal strategy. One interesting fact about this approach is that, unlike in the current reinforcement learning methods (Russell & Parr, 1995; Littman, Cassandra, & Kaelbling, 1995), it has no particular difficulty with problems where the state is only partially observable.

## 7. Future Work

To apply our work to AI planning domains such as the blocks world, we need to extend our results to richer hypothesis spaces that include first order relational predicates. There are many challenges in such domains. Concept learning from examples in such structural domains is known to be intractable (Haussler, 1989). This means that we have to extend our model to allow other kinds of information. For example, the learner might be allowed to pose its own problems to the teacher, a natural extension to the paradigm of membership queries (Angluin, 1988). Reddy et al. (1996) report a speedup learning method that learns recursive decomposition rules for planning from examples and membership queries (Reddy, Tadepalli, & Roncagliolo, 1996). A decomposition rule recursively decomposes a goal into a number of subgoals and primitive actions. As in the SIMPLEX program, the learning algorithm here needs to find a generalization of a set of positive examples. However, without the membership queries, finding a most specific generalization of a set of examples is known to be NP-hard (Haussler, 1989). The queries make it possible to find a generalization incrementally, by verifying whether each literal in the condition is relevant to the rule. We plan to extend this work to real-time domains where actions are nondeterministic and planning and execution are interleaved.

We showed that our work is applicable to the supervised setting, in which a human teacher provides solutions to problems (DeJong & Mooney, 1986; Shavlik, 1990), as well as to the unsupervised setting, where the solutions are generated by a search program (Laird et al., 1986; Minton, 1990; Tadepalli, 1992). Natarajan (1989) takes a middle course between these two extremes. The teacher is assumed to supply to the learner a set of randomly chosen "exercises" — useful subproblems that help solve the problems that naturally occur in the domain. This is very much similar to the exercises one might find at the end of a text book dealing with, say, symbolic integration or differential equations. The learner is required to converge in polynomial time and with polynomial number of exercises. Natarajan (1989) proves that the conditions sufficient for learning from solved problems as in this paper are sufficient for learning from exercises as well.

One of the challenges of unsupervised speedup learning is the "multiple image problem." In supervised speedup learning, we usually assume that the teacher's solutions are all consistent with a single problem solver in the hypothesis space. The reason that this is crucial is that every problem in the domain may have multiple solutions (images). For example, there are usually many routes to go to one's office from home, and there may be many ways of fixing a bicycle. In the absence of a teacher who ensures that all the solutions of the examples are consistent with a single problem solver, the learner has to decide if there is a target problem solver consistent with a given set of solutions or not. In other words, it has to select solutions of problems in such a way that at least one target problem solver is always retained in its current effective hypothesis space. We solved this problem in the case of Eight Puzzle by exploiting the fact that domains like Eight Puzzle have a special structure, namely serial decomposability, which allows them to have macro tables. This allowed the learner to fill any cell of the macro-table with any macro-operator that solves the corresponding subgoal, while not losing the property that the remaining macro-table can be correctly filled by other macro-operators. However, this kind of property may not hold in general in a new domain. The computational constraint that the final problem solver output by the learner must be efficient makes this particularly difficult.

Thus far we have not considered the solution quality in our analysis. In many domains like the blocks world, scheduling, and $N \times N$ generalizations of 8-puzzle, it is not difficult to find some solution to a problem, while finding optimal solutions is NP-hard (Guptha & Nau, 1992; Garey & Johnson, 1979; Ratner & Warmuth, 1986). Hence, to find reasonably good solutions in reasonable time, the learning system must make some tradeoffs. While this is achievable somewhat easily in the supervised speedup learning framework by having a helpful teacher who provides solutions that make reasonable tradeoffs, it considerably complicates the multiple image problem in unsupervised speedup learning. For example, a learner that always selects optimal solutions to the training problems, and constructs a consistent problem solver from them, may have to sacrifice efficiency of the resulting problem solver. Similarly a learner that always generalizes from easily found solutions may have to sacrifice solution quality. It is a challenging problem to design learning systems that make this tradeoff in an optimal fashion.

## 8. Conclusions

We presented a unifying formal framework to study speedup learning. Our work draws upon the extensive body of work on Probably Approximately Correct (PAC) learning. Unlike in the standard uses of PAC-learning to concept learning situations, here the learner has access to a domain-specification in terms of goals and operators, which is used to constrain the hypothesis space of problem solvers. The examples play two roles in our theory: first, they provide distribution information that determines which macro-operators or control rules are worth learning, and second, they help the learner avoid expensive search for solutions.

Our work closely integrates a number of areas which have been hitherto thought of as different − explanation-based learning, PAC learning, and empirical learning, in particular. We showed how the same framework can be used to describe learning algorithms that learn different forms of control knowledge, such as control rules and macro-operators. By introducing a single framework for speedup learning that captures seemingly dissimilar systems, we hope to have shown the underlying similarity of all these methods. In the future, we plan to extend this work to AI planning domains with richer representations and to real-time problem solving. Learning from exercises and learning to improve the quality of solutions are also worth pursuing.

## Acknowledgments

The first author is supported by the National Science Foundation under grant number IRI-9520243 and the Office of Naval Research under grant number N00014-95-1-0557. We thank Tom Amoth, Prasad Chalasani, Tom Dietterich, Oren Etzioni, Nick Flann, Sridhar Mahadevan, Steve Minton, Tom Mitchell, Barney Pell, Armand Prieditis, and Chandra Reddy for many interesting discussions on this topic. Thanks to William Cohen, Roni Khardon, Steve Minton, and the reviewers of this paper for their thorough and helpful comments, and to Padmaja Akkaraju for her careful proof-reading.